\documentclass[10pt,twocolumn,letterpaper]{article}
\pdfoutput=1
\usepackage{wacv}
\usepackage{times}
\usepackage{epsfig}
\usepackage{graphicx}
\usepackage{amsmath}
\usepackage{amssymb}
\usepackage{subfigure}
\usepackage{booktabs,multirow,multicol,ctable}
\usepackage{authblk}

\input{macros}

\newcommand{\hdrule}{\midrule[\heavyrulewidth]}

\usepackage[pagebackref=true,breaklinks=true,letterpaper=true,colorlinks,bookmarks=false]{hyperref}

\wacvfinalcopy

\def\wacvPaperID{81} 

\def\assignedStartPage{1}

\usepackage[breaklinks=true,bookmarks=false]{hyperref}
\makeatletter
\newcommand{\printfnsymbol}[1]{%
  \textsuperscript{\@fnsymbol{#1}}%
}

\begin{document}

\title{Red Carpet to Fight Club:
Partially-supervised Domain Transfer \\ for Face Recognition in Violent Videos}

\author[1]{Yunus Can Bilge\thanks{equal contribution}}
\author[1]{Mehmet Kerim Yucel\printfnsymbol{1}}
\author[2]{Ramazan Gokberk Cinbis}
\author[1]{Nazli Ikizler-Cinbis}
\author[1]{Pinar Duygulu}
\affil[1]{Department of Computer Engineering, Hacettepe University}
\affil[2]{Department of Computer Engineering, Middle East Technical University}
\affil[ ]{\textit {\{yunuscanbilge,nazli,pinar\}@cs.hacettepe.edu.tr}, mkerimyucel@hacettepe.edu.tr, gcinbis@ceng.metu.edu.tr}

\maketitle

\begin{abstract}
In many real-world problems, there is typically a large discrepancy between the characteristics of data used in training versus deployment. A prime example is the analysis of aggression videos: in a criminal incidence, typically suspects need to be identified based on their clean portrait-like photos, instead of their prior video recordings. This results in three major challenges; large domain discrepancy between violence videos and ID-photos, the lack of video examples for most individuals and limited training data availability. To mimic such scenarios, we formulate a realistic domain-transfer problem, where the goal is to transfer the recognition model trained on clean posed images to the target domain of violent videos, where training videos are available only for a subset of subjects. To this end, we introduce the ``WildestFaces'' dataset, tailored to study cross-domain recognition under a variety of adverse conditions. We divide the task of transferring a recognition model from the domain of clean images to the violent videos into two sub-problems and tackle them using (i) stacked affine-transforms for classifier-transfer, (ii) attention-driven pooling for temporal-adaptation. We additionally formulate a self-attention based model for domain-transfer. We establish a rigorous evaluation protocol for this ``clean-to-violent'' recognition task, and present a detailed analysis of the proposed dataset and the methods. Our experiments highlight the unique challenges introduced by the WildestFaces dataset and the advantages of the proposed approach.

\end{abstract}
\section{Introduction}
\label{sec:intro}

\begin{figure*}
    \begin{center}
        \centering
        \includegraphics[width=.95\textwidth]{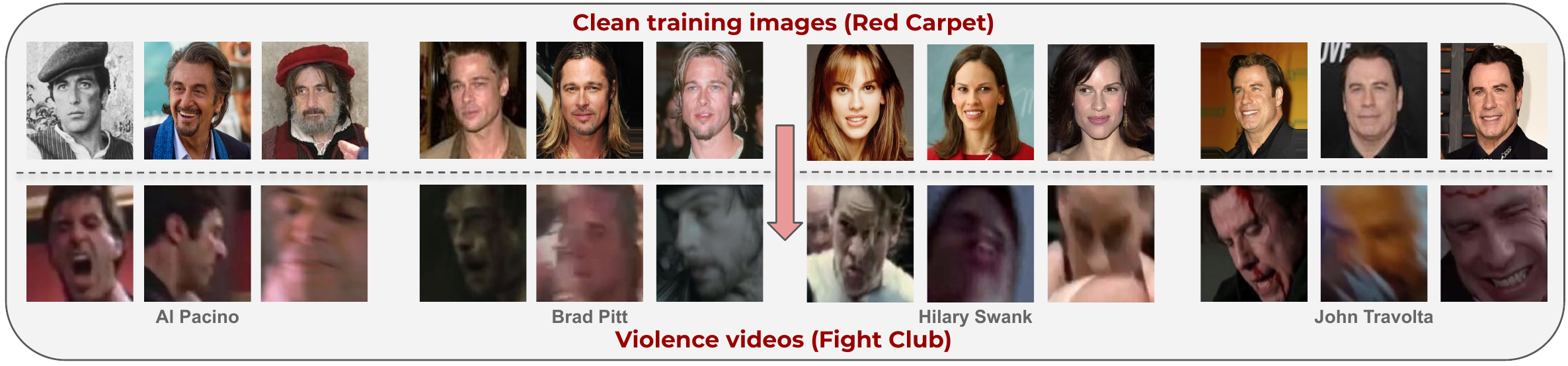}
        \caption{Our focus is the problem of transferring recognition models trained on clean, portrait-like images for person identification in the {\em wildest} (\eg fight) videos. We introduce the WildestFaces dataset that  contains videos of violent scenes from movies, together with the clean {\em red-carpet} images of the corresponding actors. Importantly, training video examples are available only for a subset of the actors, which leads to a challenging partially-supervised domain transfer problem.\label{fig:mainfigure}}
    \end{center}
    \vspace{-8mm}
\end{figure*}

People engaging in criminal activities are likely to expose a  diverse set of facial expressions/poses. The people in these activities are also
likely to move fast, causing the recorded video footage to have significant amount of blur and occlusion. What is even
more challenging is that, these people may not necessarily have prior criminal records, therefore may not have recorded
\textit{``fight scene footage's''}. They may only have \textit{``clean''} images, such as passport or Facebook-type of
photos, that can be used for identification.

In this paper, we formulate this task as transferring the face recognition model from the domain of clean (so-called \textit{Red Carpet})
images to the domain of violent (\textit{Fight Club}) videos \footnote{Our dataset is available at \url{https://ycbilge.github.io/wildestFaces}.}. Since the training videos are labeled but scarce and made available only for a subset of people, we refer to the
learning problem as {\em partially-supervised domain-transfer}.  

A plethora of studies have focused on face recognition in computer vision literature.  Compared to
the pioneering works \cite{turk1991face,ahonen2006face,xie2010fusing,edwards1998face, wright2009robust,wiskott1997face}, face recognition models that benefit from deep learning-based techniques and concentrate on better
formulation of distance metric optimization raised the bar~\cite{schroff2015facenet,taigman2014deepface,parkhi2015deep,wen2016discriminative,sun2013hybrid,sun2014deep,sun2015deepid3,deng2019arcface,zheng2018ring,wang2018cosface}. There has been interest in using additional data (in the form of unlabeled  \cite{zhan2018consensus,coelho2018face} or synthetic data \cite{zhao2017dual}), class-balancing \cite{yin2019feature}	and noisy-data handling \cite{hu2019noise} to improve face recognition accuracy. In addition to face recognition in still images,
video-based face recognition studies have also emerged (see  \cite{ding2016comprehensive} for a recent survey). Ranging
from local feature-based methods~\cite{li2013probabilistic,parkhi2014compact,li2014eigen} to manifolds
\cite{huang2015log} and metric learning~\cite{cheng2018duplex,huang2017cross,goswami2017face}, recent studies have
focused on finding informative frames in image sets \cite{goswami2014mdlface} and finding efficient ways of
feature aggregation~\cite{chowdhury2016one,yang2017neural,rao2017learning,rao2017attention}. 
Most of these studies concentrate on relatively easier cases of recognition, where the faces are seen under
good lighting conditions and are mostly stable with orthogonal viewpoints. In contrast, face recognition
in the ``wildest'' demands more than that.

In order to facilitate research in this direction, we propose a new dataset, referred to as {\em WildestFaces}. This dataset
consists of clips with adverse effects at their extreme, and auxiliary clean facial images of the corresponding people.
The videos are collected by manually finding violent scenes in movies of predetermined actors\footnote{We use {\em
actor} as a gender-neutral term, following the modern practice.}, and the clean images are collected from IMDB\footnote{www.imdb.com} and similar websites. The task is depicted in
Figure~\ref{fig:mainfigure} with example images from the dataset. 
The training set provides clean still images of
all people and violent videos of only a subset of people. The test set, however, contains novel videos of all people.
This setup resembles to that of \textit{generalized zero-shot learning (GZSL)}~\cite{xian2017zero,chao16gzsl}, where there are both seen and unseen classes in testing. We define an evaluation protocol that explicitly measures the success at recognizing people with and without training videos, and, penalizes methods that are poor at any of the two tasks.  

To tackle this {\em partially-supervised domain-transfer} problem, we divide it into two
sub-problems and propose two major techniques for each one.  First, we leverage stacked affine-transform 
layers for classifier transfer, which aims to adapt the image-based classification model to the target-domain
representations, in a supervised yet data-efficient manner. Second, we propose an attention-driven temporal pooling
layer, which aims to enable data-driven adaptation to the face tracks in the video domain. We additionally propose a
third self-attention~\cite{vaswani2017attention} based formulation, with components targeted to both sub-problems.  We rigorously evaluate all
proposed techniques and show their advantages over a number of state-of-the-art alternatives.

To sum up, the contributions of this paper include: 
{\bf (1)} A new dataset called WildestFaces that includes a wide range of examples from violent movies;
{\bf (2)} A new partially-supervised face recognition task, where classifiers trained on clean image data are evaluated for their ability of recognizing faces in violent videos;
{\bf (3)} Rigorous evaluation protocols inspired from the recent developments in related problems (primarily zero-shot learning and few-shot learning);
and {\bf (4)} Effective techniques for the proposed partially-supervised, \textit{clean-to-violent} domain-transfer problem.

\section{Related Work} \label{related_work}

\noindent \textbf{Face Recognition Datasets:} Due to the data-hungry nature of face recognition, there have been many attempts in building large scale datasets. FDDB \cite{jain2010fddb}, AFW \cite{zhu2012face}, PASCAL Faces \cite{yan2014face}, Labeled Faces in the Wild (LFW) \cite{huang2007labeled}, Celeb Faces \cite{sun2013hybrid}, Youtube Faces  (YTF) \cite{wolf2011face}, FaceScrub \cite{ng2014data}, IJB-A \cite{klare2015pushing}, MS-Celeb-1M \cite{guo2016ms}, VGG-Face \cite{parkhi2015deep},  VGG2-Face \cite{cao2017vggface2}, MegaFace \cite{kemelmacher2016megaface} and WIDER Face \cite{yang2016wider} datasets have been made publicly available for research purposes. Datasets with extreme scales, such as \cite{schroff2015facenet} and \cite{taigman2014deepface} have also been used but have not been disclosed to the public. 

For video face recognition, YouTube Faces \cite{wolf2011face} is the most widely used dataset. While it contains motion-blurred and low-quality frames,
overall the quality of the frames is typically much better than the wildest conditions that we target in our work. Plus, unlike our domain-transfer based 
image-to-video recognition setup, in YouTube Faces, the primary focus on video-to-video recognition. Other two prominent video face recognition datasets are COX \cite{huang2015benchmark} and PasC \cite{beveridge2013challenge}. Despite their relatively large size, PasC \cite{beveridge2013challenge} suffers from video location constraints and COX \cite{huang2015benchmark} suffers from demographics as well as video location constraints. FaceScrub \cite{ng2014data} is a dataset which has resemblance to our case as it also includes actors as individuals. However the dataset only contains actor images rather than videos. \textcolor{black}{The most relevant benchmarks to ours are \cite{kalka2018ijb,ferrari2018extended, sengupta2016frontal, kushwaha2018disguised}. However, none of them specifically focus on domain shifts,  recognizing \textit{unseen} classes (from a ZSL perspective) or violent settings.}

\noindent \textbf{Video Face Recognition:}  
\cite{yang2017neural} employ attention modules to adaptively aggregate image-based features from frames into a single representation.  Instead of aggregating feature representations, \cite{rao2017learning} opt to aggregate raw frames directly to produce a synthesized image via generative models that is tailored to be more discriminative. \cite{tran2017disentangled} utilizes an encoder-decoder structure in their GAN's generator to achieve a pose-independent identity representation, which is later used to synthesize an image of desired pose. A similar work exploiting attention-like mechanism is presented in \cite{xie2018multicolumn}, where content and visual quality of each image is learned to perform set-wise classification. \cite{rao2017attention} exploits reinforcement learning to attend to informative frames in videos which are aggregated by a mean-pooling to represent sets as a single feature vector. \cite{wu2018light}, on the other hand, presents a light-weight network to achieve fast face recognition. \cite{zhao2017dual} employ video-domain only face recognition technique in a fully supervised manner. The method is tailored for single-domain face recognition and for seen classes. In template-based face recognition, a similar effort to produce a single representation is given in \cite{zhong2018ghostvlad}. These papers, however, do not address the explicit domain-transfer problem in \textit{clean-to-violent} face recognition problem and operate in the domains of the datasets they are trained on.

\textcolor{black}{A recent work that is most closely related to ours is \cite{nagrani2018benedict}, where the main goal is to identify characters in TV series videos using classifiers trained on clean actor images. Their main focus is different in many ways from ours since \cite{nagrani2018benedict} (i) presumes that a large number of weakly supervised video examples are available for all characters, (ii) uses voice classifiers and shows that identification is largely influenced by them, and, (iii) leverages similarities across different scenes within a single dataset during recognition.}

\noindent \textbf{Domain Adaptation:}  
Due to its dual-domain nature, clean-to-violent face recognition poses a domain-transfer problem. Having found application areas in primarily computer vision tasks~\cite{wulfmeier2017addressing,wu2017compact,wang2017deep}, supervised~\cite{chen2017show,gebru2017fine,tzeng2015simultaneous} and unsupervised variants~\cite{ganin2016domain,ghifary2016deep,yan2017mind, long2016unsupervised} of domain adaptation techniques have surfaced in recent years. There are several approaches pertinent to the task, such as feature space alignment \cite{sun2016return}, supervised feature transformation~\cite{csurka2016unsupervised,long2014adaptation}, adversarial approaches \cite{goodfellow2014generative}, encoder-decoder structures ~\cite{bousmalis2016domain,ghifary2016deep} and many others. \textcolor{black}{For a detailed review in domain adaptation, readers are referred to \cite{csurka2017domain}.}

The main difference between the mainstream domain-adaptation tasks and our problem definition is that in domain adaptation, it is typically presumed that
(labeled or unlabeled) target-domain training examples are available for all classes of interest, which is not realistic for our clean-to-violent recognition problem. To this end, our work aims to (i) address a partially supervised domain-transfer problem, (ii) handle unseen class recognition, (ii) introduce evaluation protocols for partially-supervised transfer and (iii) learn to handle noisy sequences. A similar work to ours in domain adaptation literature is \cite{long2016unsupervised}, however ours differ from this work by the factors listed above and ours is also geared towards dual-domain (image-to-video and clean-to-Wildest) face recognition.

\noindent \textbf{Zero-shot learning:}
In zero-shot learning (ZSL), the classifiers are learned over seen classes and then extended to unseen classes of which labeled data is not accessible, by means of auxiliary data such as attributes or textual descriptions. Generalized zero-shot learning (GZSL)~\cite{xian2017zero,chao16gzsl} extends the test protocol of ZSL to include seen and unseen classes together, as it is more natural to assume cooccurence of these classes in general. 
Different from regular GZSL, the auxiliary information is in the form of a set of labelled images, as opposed to attributes or textual descriptions as mostly used in the mainstream zero-shot learning research.  

Overall, the proposed problem setup is at the intersection of GZSL and supervised domain adaptation, where training video data is available only for a subset of classes.

\section{WildestFaces Dataset} 
\label{dataset_info}

To the best of our knowledge, there is no publicly available dataset which is composed of fight and dispute videos, with annotated human faces. We introduce the \textit{WildestFaces} dataset collected by focusing on violent movie scenes of celebrities. Below, we give the details of the dataset and the collection procedure.

\noindent {\bf Videos.} We first created a list of actors appearing in movies with violence. We then collected videos of them from YouTube using a variety of scene settings; \eg car chase, fist fights, gun fights, heated arguments, \etc. This abundance in scene settings provide an inherent variety of occlusions, poses, background clutter and motion blur. Videos, with an average 25 FPS are then divided into shots with a maximum duration of 10 seconds. 

In total, for 64 selected actors, 2,186 shots (64,242 frames) from 410 videos are collected. We annotated the face regions by applying a face detector and manually correcting its mistakes.  In this process, no frames were filtered out due to adverse conditions; and we labeled even extremely tiny, occluded, frontal/profile and blurred faces. 

\noindent {\bf Clean images.} In order to employ classifier transfer from clean images to violence videos, we also collected images of actors taken under normal conditions such as red carpet images. We primarily use IMDB-WIKI \cite{RotheIJCV2016}, from which we acquire the images of 62 celebrities that overlap with the video subjects. For the remaining two subjects, we collect images from the Internet. In total, we obtain 8069 images of 64 subjects. A detailed analysis of the dataset is presented in Section~\ref{results}.

\section{Partially-supervised domain-transfer} 
\label{sec:method}
\label{sec:method:probldef}

\newcommand{\R}{\mathbb{R}}
\newcommand{\X}{\mathcal{X}} 
\newcommand{\V}{\mathcal{V}} 
\newcommand{\Dx}{\mathcal{D}_x} 
\newcommand{\Dv}{\mathcal{D}_v} 
\newcommand{\Y}{\mathcal{Y}} 
\newcommand{\Ys}{\mathcal{Y}_\text{seen}} 
\newcommand{\Yu}{\mathcal{Y}_\text{unseen}} 
\newcommand{\T}{{\sf T}} 

We assume that during training, there are two sets:
(i) the source domain training set $\Dx={(x_i,y_i)}_{i=1}^{n_x}$ with $n_x$ still image examples, and, (ii) the 
target video domain training set $\Dv={(v_j,y_j)}_{j=1}^{n_v}$ with $n_v$ video examples. Each example $(x_i,y_i)$
in the image training set $\Dx$ contains the facial image $x_i \in \X$ and the corresponding person label $y_i$.
Each example $(v_j,y_j)$ in the video training set $\Dv$ contains the facial image sequence $v_j \in \V$ of length $|v_j|$, with
frames denoted as $v_j = (v_j[1],...,v_j[|v_j|])$, and the label $y_j$.

The crucial detail is the difference between set of classes spanned by these two training resources: while $\Dx$ provides examples for the set $\Y$ of all classes, $\Dv$ provides examples
only for a subset of them. However, at test time, an input video may belong to any class in $\Y$. Inspired from the
similarity to the generalized zero-shot learning problem (as discussed in Section~\ref{related_work}), we refer to the set of classes having
both image and video training examples as the {\em seen} classes and denote them by $\Ys$, and, the remaining set of 
classes having only image domain examples as the {\em unseen} classes and denote them by $\Yu$. We denote the number of seen and unseen classes by $c_s$ and $c_u$, respectively \footnote{We adapted the GZSL nomenclature, as clean images and training videos are akin to class descriptions and seen class examples, respectively.}. 
The final goal is to learn a classifier scoring function $f_v: \V \rightarrow \R^{|\Y|}$ that maps a video-domain input
to the vector of per-class confidence scores for all classes. We divide this task in two sub-problems and propose methods towards each one 
in the following sections: (i) classifier transfer, (ii) temporal adaptation. We additionally propose a third self-attention~\cite{vaswani2017attention} based approach that 
aims to tackle both sub-problems.

\subsection{Classifier Transfer}
\label{sec:method:modeladapt}

Let $\phi: \X \rightarrow \R^{d_x}$ be an image-domain feature extractor that maps each input face image to a
$d_x$-dimensional vector, and, let $\Psi: \V \rightarrow \R^{d_v}$ be a video-domain feature extractor that maps each
input face video to a $d_v$-dimensional vector. Throughout our experiments, we use a pre-trained VGG-face network~\cite{parkhi2015deep} as
$\phi$ (Section~\ref{results}). We propose a number of $\Psi$ alternatives in Section~\ref{sec:method:temporal}.
In classifier transfer, the goal is to adapt a classifier pre-trained in the source domain using $\phi$ representation,
to the target domain with representation $\Psi$ via the restricted set of examples for the $c_s$ classes.

\newcommand{\Wcls}{{W}} 
\newcommand{\wcls}{{w}} 

We start by defining the source domain classifier. For simplicity, we use a linear model for source-domain classification,
parameterized by the matrix $\Wcls = [\wcls_1,...,\wcls_{|\Y|}] \in \R^{d_x \times |\Y|}$.
The model is trained on the source domain dataset $\Dx$ via regularized loss minimization:
\begin{equation}
    \min_{W} R(\Wcls) + \sum_{i=1}^{n_x} \ell( \phi(x_i)^\T \Wcls, y_i ) 
\end{equation}
where $\ell(\cdot,y)$ is the soft-max cross-entropy loss function, and, $R(\Wcls)$ is $\ell_2$-regularization in our 
experiments.\footnote{While we exclude the regularization weight from the equations for brevity, we tune it on the validation set and utilize in our experiments.}

We formalize the classifier-transfer problem as the task of learning a transformation $\tau: \R^{d_x} \rightarrow
\R^{d_v}$ by minimizing the regularized loss on the target-domain dataset $\Dv$:
\begin{equation}
    \min_{\tau} R(\tau) + \sum_{j=1}^{n_v} \ell( \tau(\Psi(v_j))^\T \Wcls, y_j ) 
\label{eq:clstrans}\end{equation}
where $R(\tau)$ represents the regularization applied to the transfer model. In this framework, $\tau$ has the
responsibility of transforming the input $\Psi(v)$ video-domain representation to a $\phi(x)$-like image-domain
representation and make it compatible with the classification layer $\Wcls$.  When training on the $\Dv$ dataset, we
deliberately keep the classification layer $\Wcls$ fixed, in order to keep the class models $\wcls_j$ intact and
compatible with each other, and, minimize the risk of learning a bias towards the subset of classes seen in $\Dv$.

The first classifier-transfer technique that we consider is the {\bf fully-connected classifier-transfer} layer:
\begin{equation}
    \tau_\text{fc}(\Psi(v)) = Q \Psi(v) ,
\end{equation}
where $\tau_\text{fc}$ is instantiated by a fully-connected layer (fc) $Q \in \R^{d_x \times d_v}$, and,
linearly transforms the $d_x$ dimensional image representation to the $d_v$ dimensional face-track vector.

While the aforementioned approach looks simple and promising, it performs poorly in practice: the number of parameters
in $Q$ is typically too high to be trained properly unless the target-domain dataset $\Dv$ is
large-scale, which is infeasible in most practical scenarios, including ours. For instance, when VGG-face $\phi$ descriptors
are being used and $\Psi$ is defined as the average of per-frame VGG-face descriptors, $Q$ contains $4096^2$
($\sim16$M) parameters. As a result, it quickly leads to over-fitting, and yields a poor trade-off between the
seen and unseen class performance in the target-domain (Section~\ref{results}).

In our preliminary experiments, we have investigated a number of common regularization techniques, including $\ell_2$,
drop-out, batch-normalization and explicit rank regularization, and in all cases, we have observed very similar poor
generalization behavior for the fc based classifier transfer. 

To avoid these difficulties, we propose the {\bf affine classifier-transfer} layer, which is built upon a much
lower-complexity affine model:
\begin{equation}
    \tau_\text{affine}(\Psi(v)) = \alpha \odot \Psi(v) + \beta,
\end{equation}
which implements feature scaling via applying Hadamard Product with the vector $\alpha$, and, shifting by the vector $\beta$, which
are trained according to Eq.~\ref{eq:clstrans}.  The underlying assumption here is
that the source-domain and target-domain representations are of the same dimensionality and are
sufficiently correlated so that an affine transform can provide the necessary correction. Fortunately, this assumption
is met in most practical $\Psi$ definitions, including temporal average pooling and attentive temporal pooling, which are explained
in the following section. 

We propose two extensions to the affine classifier-transfer layer. First, we propose the {\bf stacked affine classifier-transfer} where 
the affine transform is followed by the ReLU activation and then another affine transform. Second, we propose the {\bf residual stacked affine classifier-transfer} (rsa) layer, which includes a residual connection:
\begin{equation}
    \tau_\text{rsa}(\Psi(v)) = \alpha_2 \odot \max(\alpha_1 \odot \Psi(v) + \beta_1,0) + \beta_2 + \Psi(v)
\end{equation}
where $(\alpha_1,\beta_1)$ and $(\alpha_2,\beta_2)$ are the parameters of the first and the second affine transforms, respectively.
In Section~\ref{results}, we thoroughly evaluate all major cases of classifier-transfer and the combinations with temporal adaptation techniques.

\begin{figure*}[!ht]%
\begin{center}
\vspace{-6mm}
\includegraphics[width=\textwidth]{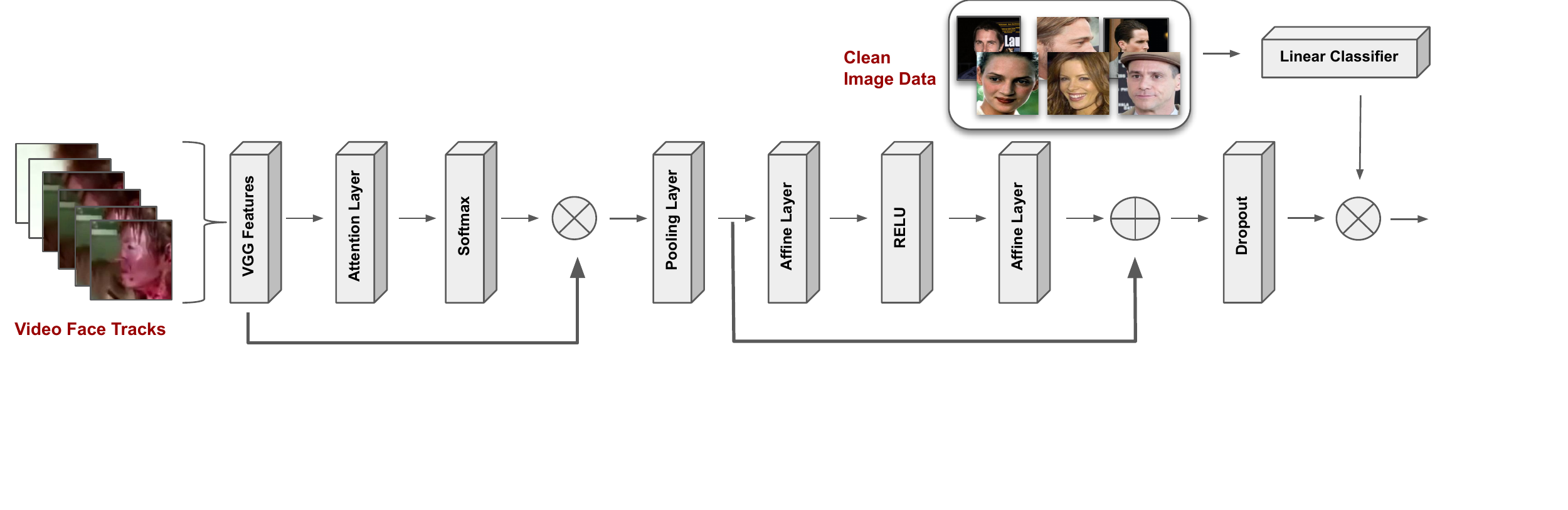}
\end{center}
\vspace{-2cm}
    \caption{The architecture of the proposed domain transfer approach based on Attentive Temporal Pooling (ATP) layer for temporal-adaptation, combined with residual stacked affine classifier-transfer layer.}
\label{methodFig}
\vspace{-5mm}
\end{figure*}

\subsection{Temporal Adaptation}
\label{sec:method:temporal}

We now continue within the framework defined in the previous section, and propose techniques 
for obtaining a video representation $\Psi(v)$ suitable for the clean-to-violent domain-transfer task.

For clarity, we start with the simple {\bf temporal average pooling} scheme. In this case, the representation of a face
track is obtained by taking average of the per-frame features extracted using the image-domain feature
extractor $\phi$:
\begin{equation}
    \Psi_\text{AvgPool}(v) = \frac{1}{|v|} \sum_{t=1}^{|v|} \phi(v[t])
\end{equation}
While temporal average pooling is a versatile technique, the resulting representation is likely to be dominated by the
heavily motion-blurred faces in a track, plus, it is likely to handle multiple poses poorly.  Similarly, temporal max
pooling, an obvious alternative, is likely to be negatively affected by these factors.\footnote{In fact, we empirically
observe that max-pooling performs similar to or worse than average-pooling except when self-attention is being used. We
do not report simple max-pooling results in Section~\ref{results} for brevity.}

To handle the multi-modality and noise in face tracks, we aim to learn a data-driven temporal representation optimized for the
clean-to-violent domain-transfer task.  For this purpose, we propose {\bf attentive temporal pooling} (ATP), inspired
from \cite{yang2017neural}. The intuition behind this model is to exploit the hidden pose information in a trainable
fashion \textcolor{black}{(unlike other pooling strategies which require additional input or manual invertention \cite{hassner2016pooling, ferrari2019discovering})} to extract useful information in the noisy sequences of video frames. The proposed approach consists of two
main components: (i) an attention layer, and, (ii) a attention-weighted pooling layer. Attention module
learns to promote the informative parts of given image sequences. Through the pooling layer, the overall sequence
information is aggregated.

More formally, assuming that per-frame descriptors are extracted using $\phi$, we define the attention weight matrix
$A=[a_1,...,a_K] \in \R^{d_x \times K}$, where $K$ can be interpreted as the hyper-parameter defining the number of
canonical appearance modes used in the attention model.  The attention function $\Gamma(v)$ computes a $|v| \times K$
attention matrix, whose $t$-th frame, $k$-th mode value is given by applying a temporal softmax over the raw attention
scores:
\begin{equation}
    [\Gamma(v)]_{t,k} = \frac{\exp\left[ \phi(v[t])^\T a_k \right]}{\sum_{t^\prime=1}^{|v|} \exp\left[\phi(v[t^\prime])^\T a_k \right]} .
\end{equation}
The $k$-th column of the resulting matrix can be considered as a weight distribution over the frames. We use these 
weights in temporal pooling to obtain $K$ different representations, \ie a $d_x \times K$ dimensional
matrix given by $\Phi(v) \Gamma(v)$, where $\Phi(v)=[\phi(v[t])]_{t=1}^{|v|} \in \R^{d_x \times |v|}$ is the matrix
of all per-frame $\phi$ representations of the facial video $v$.  These per-mode descriptors are then aggregated
into a single $d_x$-dimensional vector using average-pooling (Figure~\ref{methodFig}). The overall operation can equivalently be expressed as:
\begin{equation}
    \Psi_\text{ATP}(v) = \frac{1}{K} \Phi(v) \Gamma(v) \mathbf{1}_{K}
\label{eq:PsiATP}
\end{equation}
where $\mathbf{1}_{K}$ is the $K$-dimensional vector of all ones. This expression also reveals that the ATP 
scheme effectively assigns an attention weight to each frame, where all unnormalized per-frame weights are given by $\Gamma(v) \mathbf{1}_{K}$.

Using the previously defined domain-transfer framework, we learn the ATP model jointly with the classifier-transfer
model $\tau$ on the dataset $\Dv$:
\begin{equation}
    \min_{\tau,\Psi_\text{ATP}} R(\tau) + R(\Psi_\text{ATP}) + \sum_{j=1}^{n_v} \ell( \tau(\Psi_\text{ATP}(v_j))^\T \Wcls, y_j ) 
\label{eq:TrainATP}
\end{equation}
which corresponds to learning a data-driven temporal representation for the domain-transfer task.

\subsection{Self-attention based domain-transfer}

In addition to the techniques that we propose for classifier-transfer and temporal-adaptation, we define \textcolor{black}{another baseline method}, 
a self-attention \cite{vaswani2017attention} based formulation that aims to jointly tackle both sub-problems.
Self-attention mechanism aims to capture the internal structure of a sequence by learning the inter-element relations.
Below we briefly explain the way we adapt it to the domain-transfer problem, and, refer 
to \cite{vaswani2017attention} for a full specification of the original approach.

A self-attention layer consists of three transforms for computing the {\em key}, {\em query} and {\em value} tensors for
each element.  The attention weight of each element (\ie face) in a sequence \wrt each other element is computed based on
the per-element query and key embeddings, and the attention-driven representation of each element is obtained by computing
the attention-weighted sum of all per-element value embeddings. In this sense, self-attention has certain similarities
to ATP (Eq.~\ref{eq:PsiATP}), with two major differences: (i) while ATP learns $K$ canonical attention-references,
self-attention uses each element within a sequence as an attention-reference on its own, (ii) while ATP yields a
weighted summation of original per-frame descriptors, self-attention additionally learns a descriptor transformation,
which can be considered as the classifier-transfer layer.

In the context of our domain-transfer problem, we have observed that it is beneficial to (i) carefully tune the output dimensionality of
key and query transforms on the validation set, (ii) utilize {\em position-wise} feed-forward network component~\cite{vaswani2017attention},
(iii) use max-pooling (instead of average-pooling) to aggregate the final per-element embeddings to a single vector.
We set the value dimensionality to $d_x$, so that the classifier layer can be applied to the
resulting video representation $\Psi$. We learn the parameters of the network on $\Dv$ pretty much the 
same way as in ATP training (Eq.~\ref{eq:TrainATP}).

\section{Experimental Results} \label{results}

In this section, we present a detailed analysis of the WildestFaces dataset, the experimental setup, evaluation protocols, and, the evaluation of the proposed approaches.

\subsection{Dataset analysis} 

\begin{figure}[t]%
\begin{center}
\subfigure{%
			\includegraphics[width=0.06\textwidth]{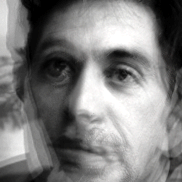}%
			\includegraphics[width=0.06\textwidth]{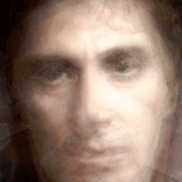}%
			\includegraphics[width=0.06\textwidth]{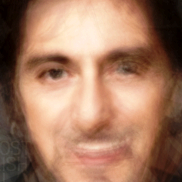}%
			\includegraphics[width=0.06\textwidth]{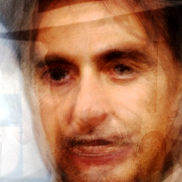}%
			\includegraphics[width=0.06\textwidth]{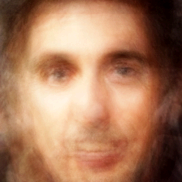}%
			\includegraphics[width=0.06\textwidth]{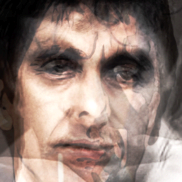}%
			\includegraphics[width=0.06\textwidth]{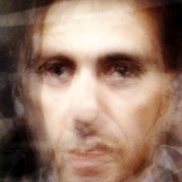}%
			\includegraphics[width=0.06\textwidth]{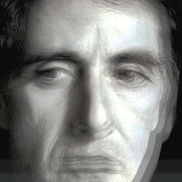} 
			}\\
			\vspace{-2mm}
		\subfigure{%
			\includegraphics[width=0.06\textwidth]{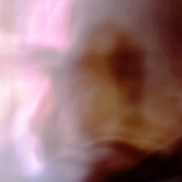}%
			\includegraphics[width=0.06\textwidth]{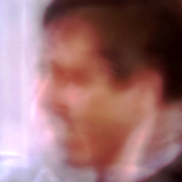}%
			\includegraphics[width=0.06\textwidth]{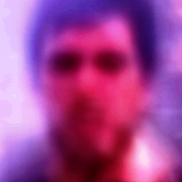}%
			\includegraphics[width=0.06\textwidth]{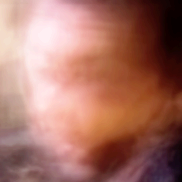}%
			\includegraphics[width=0.06\textwidth]{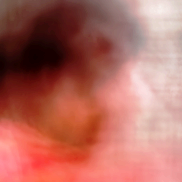}%
			\includegraphics[width=0.06\textwidth]{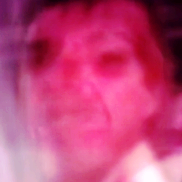}%
			\includegraphics[width=0.06\textwidth]{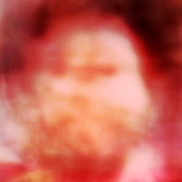}%
			\includegraphics[width=0.06\textwidth]{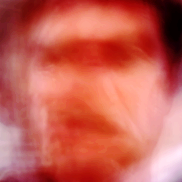}
			} 
			
\end{center}
\vspace{-3mm}
\caption{K-Means cluster centers (with $k=8$  for Al Pacino images in FaceScrub \cite{ng2014data} and \textit{WildestFaces} datasets are shown in first and second row. Average faces from \textit{WildestFaces} are hardly recognizable, indicating a large degree of variance in adverse effects. Images are histogram equalized for convenience. Better viewed when zoomed in.} \label{face_clusters}
\vspace{-5mm}
\end{figure}

In Figure~\ref{face_clusters}, k-means centers of all Al Pacino images are shown for FaceScrub \cite{ng2014data} and WildestFaces datasets. It can be seen that WildestFaces has a wide spectrum of adverse effects as its cluster centers are not recognizable. WildestFaces offers a good distribution of blur levels, pose variance, and a noticeable age variance, where approximately half of all shots are occluded. Dataset splits are summarized in Table~\ref{tab:protocols}. Below we present the detailed statistics.
\vspace{2mm}

\begin{figure*}
	\begin{center}
		\subfigure[Scale statistics.]{%
			\label{scale_rec}%
			\includegraphics[width=0.23\linewidth]{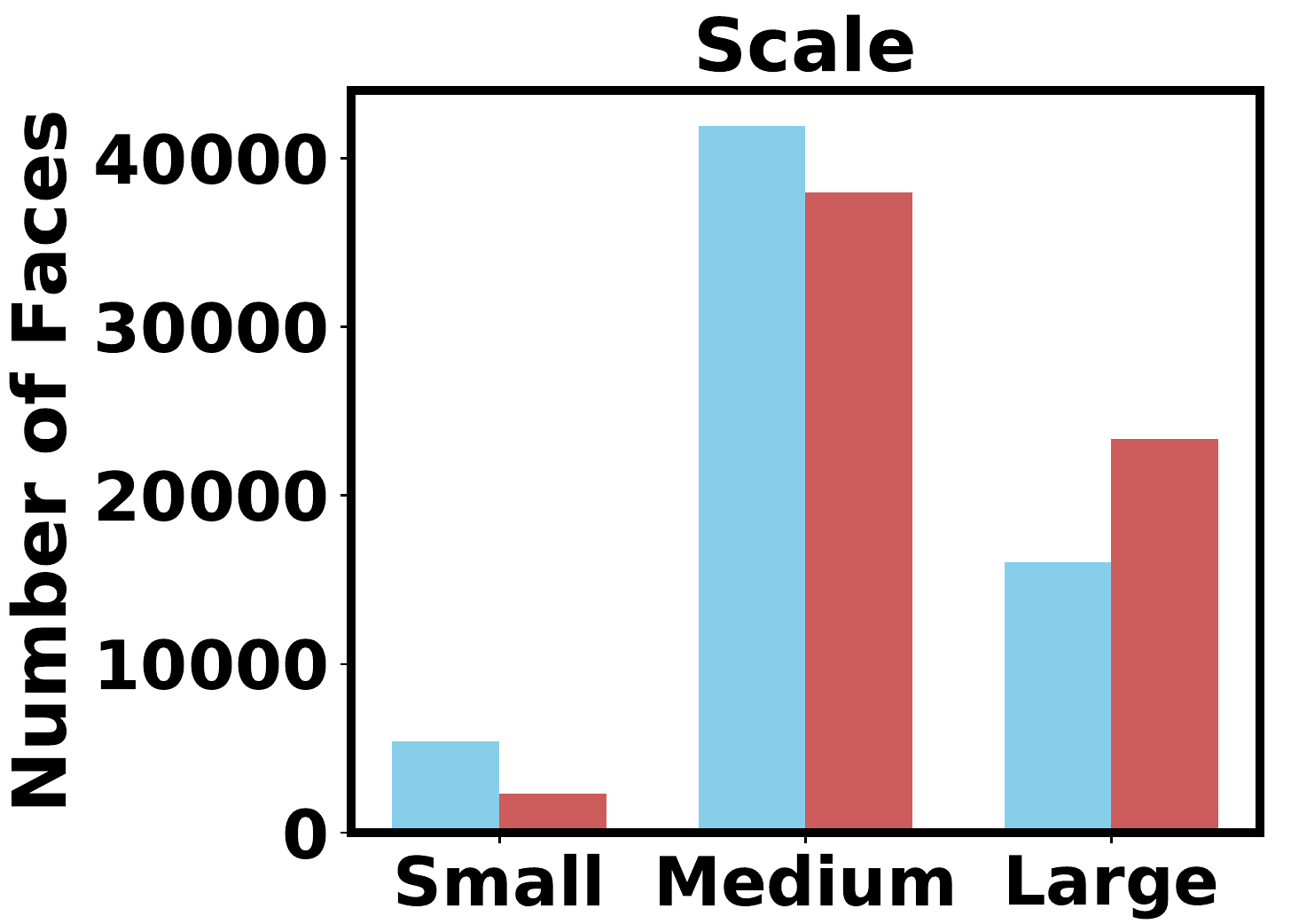}}
		\subfigure[Blur statistics.]{%
			\label{blur_rec}%
			\includegraphics[width=0.23\linewidth]{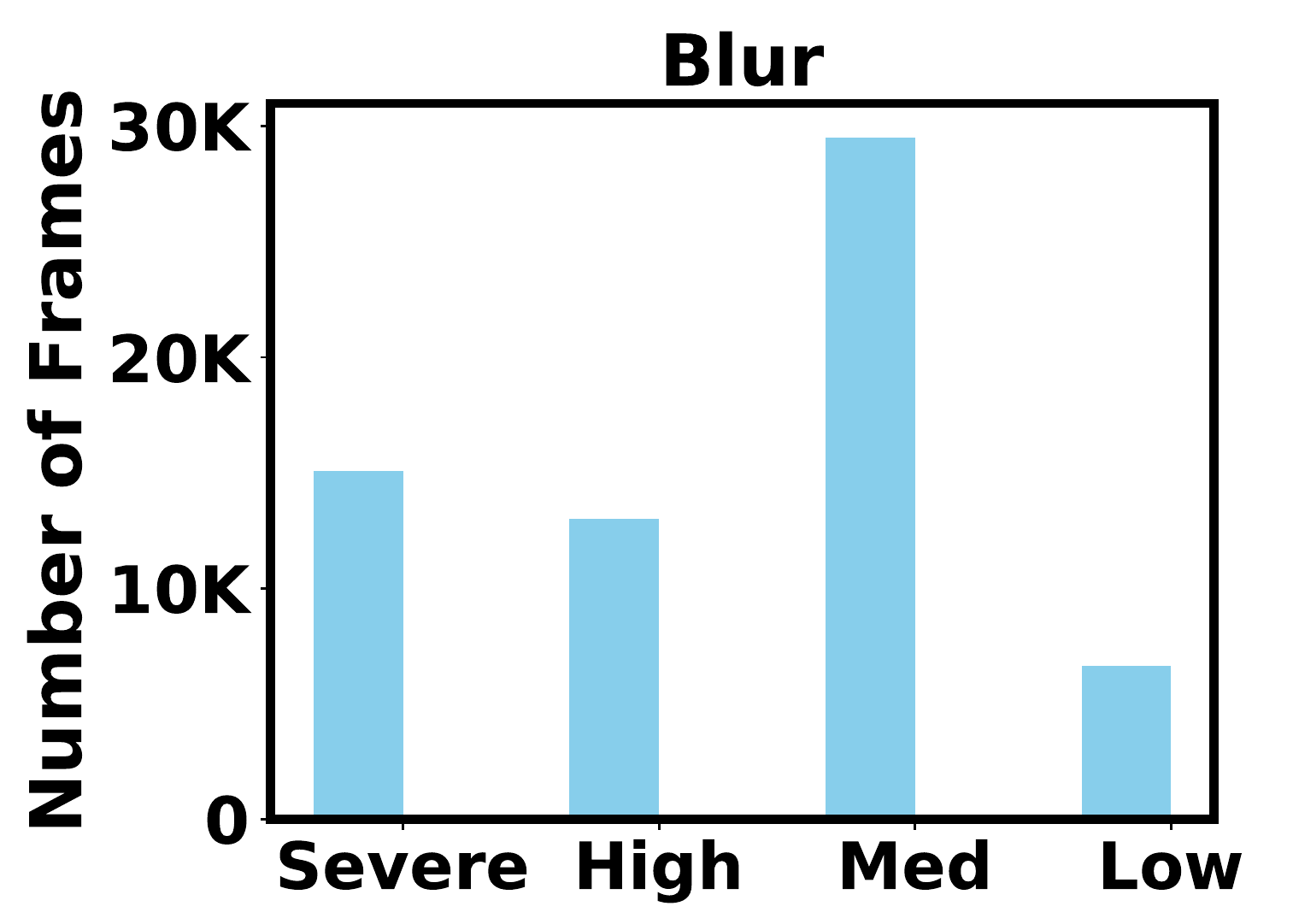}}%
		\subfigure[Age variance.]{%
			\label{age_variance}%
			\includegraphics[width=0.22\linewidth]{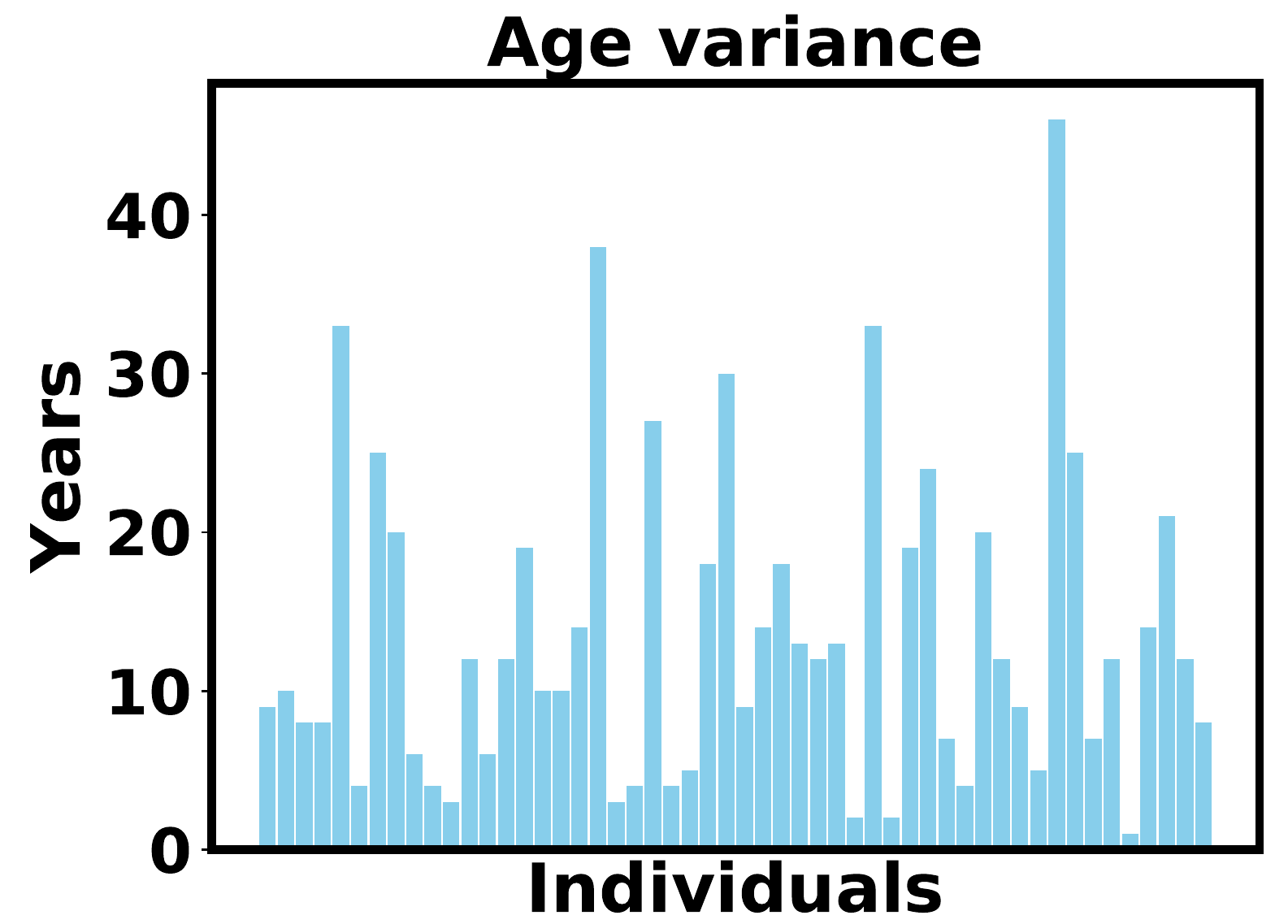}}%
		\subfigure[Pose statistics.]{%
			\label{pose_variance}%
			\includegraphics[width=0.23\linewidth]{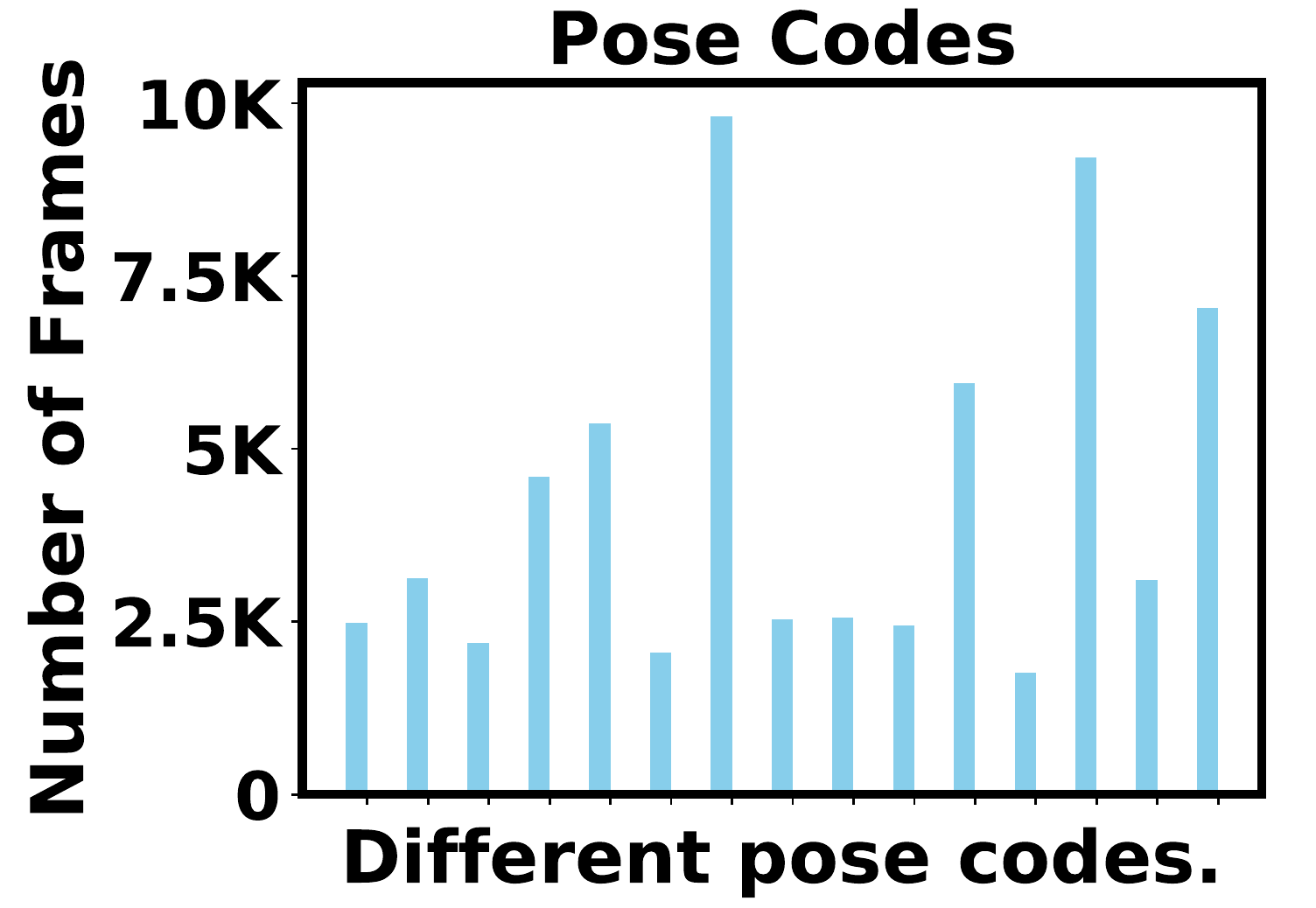}}%
			\vspace{-2mm}
                \caption{\textit{WildestFaces} Statistics. (a) Blue  and  red  correspond  to  width  and  height, respectively. (b) Blur statistics indicate an emphasis on medium blur. (c) and (d) show that the pose and age variances are high. }
			\vspace{-3mm}		
	\end{center}
        \label{fig:stats}
        \vspace{-3mm}
\end{figure*}

\noindent\textbf{Scale.} Faces below 100 pixels are accounted as \textit{small}, in between 100 to 300 pixels as \textit{medium}, and larger than 300 pixels as \textit{large}. Scale statistics given in Figure~\ref{scale_rec} shows that \textit{medium} size is more common.

\noindent\textbf{Blur.} 
Inspired from \cite{pech2000diatom}, we perform contrast normalization and grayscale conversion. These images are convolved with a 3x3 Laplacian Kernel, and variance of the result is used to produce a blurness value. Blur values are used to empirically find a threshold to categorize images in blur levels. Blur statistics are shown in Figure~\ref{blur_rec}. 

\noindent\textbf{Age.} For each individual, we measure the differences between the dates of their earliest and latest movies. We observe age variations up to 40 years (Figure~\ref{age_variance}), where the average variation is 13 years.

\noindent\textbf{Occlusion.} We randomly select $250$ shots and label them according to the amount of occlusion present. We observe that 20\% have \textit{no occlusion}, 28\% have \textit{medium} and 52\% have \textit{significant} occlusion.

\noindent\textbf{Pose.} We use \cite{ruiz2017fine} to find orientations of the faces and then quantize them using k-means to find the pose codes. Figure~\ref{pose_variance} shows the distribution of various pose codes. We observe that pose variance is a major challenge.

\subsection{Experimental setup}

Supervised evaluation is not fully realistic in our case; not every individual may have a criminal record history and corresponding fight or dispute video footage(s). The only available means of identification can be clean images. Our evaluation protocol mimics this scenario, where the test set includes videos of individuals that are not seen before. 

\noindent \textbf{Evaluation protocol and metrics.}  Training, test and validation sets for WildestFaces are split person-wise,  \textcolor{black}{where classes with fewer per-class sample counts are selected as \textit{unseen} classes}. For the partially-supervised domain-transfer evaluation, inspired
from the recently developed evaluation protocols for generalized zero-shot learning~\cite{xian2017zero,chao16gzsl} and
generalized few-shot learning~\cite{gidaris18gfew}, we use the following protocol and metrics: the recognition model has
access to the still image training examples of all 64 classes, the training videos of 40 classes, and the validation
videos of additional 10 persons. At test time, an input image may belong to any one of the 64 classes. The
normalized accuracy values over the test images of seen and unseen classes are averaged separately. The final performance score
is defined as the harmonic mean (test-h) of the seen and unseen class accuracies.
The same procedure is also utilized to obtain validation set harmonic mean score (val-h).

\noindent \textbf{Implementation details.}  We use VGGFace \cite{parkhi2015deep} for image representation.  We tune all the hyper-parameters based on the
val-h metric and decide whether to use dropout or not for each method separately. We use early stopping to introduce further regularization. While training, we take samples from each class with probability inversely proportional to its number of training examples to deal with class imbalance. The models are implemented in PyTorch \cite{paszke2017pytorch}, parameters are initialized using Xavier~\cite{glorot2010understanding}, and the results are averaged over 10 runs to mitigate the occasional fluctuations.

\subsection{Results and discussion}

\begin{table}
\centering
\caption{Dataset splits. \textcolor{black}{The upper two rows and the bottom one correspond to WildestFaces and IMDB datasets, respectively.} }
\label{tab:protocols}
{\small
\begin{tabular}{lccc}
\toprule
&  Train & Validation &  Test \\ 
\hdrule
Shots & 1156 & 495 & 535 \\ 
Images  & 35051 & 14520  &  14671\\
\cmidrule{1-4}
Images & 6428 & n/a & 1641 \\ 
\bottomrule
\end{tabular}
}
\vspace{-6mm}
\end{table}

\subsubsection{Classifier Transfer}
We evaluate the effectiveness of fc and affine transform based classifier transfer methods.
We consider two main domain-adaptation baselines, based on MMD~\cite{Long:2015:LTF:3045118.3045130} and the adversarial training method of ADDA~\cite{tzeng2017adversarial}. For both, we consider the fixed VGG feature extractor as the source domain mapping and 
aim to learn a target-to-source mapping that can transform the video representation to the source domain.
We define their fc and affine transform based versions, which yields four domain adaptation baselines.

Results are shown in Table \ref{tab:classifiertransfer}. 
Fully-connected classifier layer (fc) performs poorly, due to heavy overfitting as a result of having a
large number of trainable parameters, despite our tuning efforts. While MMD-affine improves over MMD-fc, neither method improves over results w/o any transfer layer. ADDA-fc fails to converge even with one fc layer (not shown for brevity). ADDA-affine, on the other hand, proves effective and improves from 27.3 to 31.7. The proposed affine transfer further improves to 35.4. We also train MMD-affine and ADDA-affine baselines with train and validation videos, but observe only neglibible improvements despite adding examples from 10 new classes.

\begin{table}
\begin{center}
\caption{Comparison of classifier-transfer methods (with one affine layer and temporal average pooling).} 
\label{tab:classifiertransfer}
\begin{tabular}{lcccc} 
\toprule
                 & seen & unseen & harmonic \\
\midrule
Random 		     & 2.5          & 4.1              & 3.1 \\
No transfer     & 29.3         & 25.5           & 27.3 \\
Fully-connected               & 20.2       & 7.2            & 10.3 \\
MMD-fc \cite{Long:2015:LTF:3045118.3045130}             & 25.8         &  22.1          & 23.6 \\
MMD-affine       &  28.0        &  23.2             & 25.2\\
ADDA-affine \cite{tzeng2017adversarial} & 35.2 & 29.3 & 31.7 \\
\midrule
Affine (Ours)    & \textbf{39.6}      & \textbf{32.2}        & \textbf{35.4} \\
\bottomrule
\end{tabular}
\vspace{-8mm}
\end{center}
\end{table}

In Table \ref{avgpoolvsatp}, we experiment with the stacked affine classifier-transfer, and residual stacked classifier-transfer layers together with AvgPool and ATP temporal adaptation techniques. As can be seen, amongst different variations, 2-layer residual stacked classifier-transfer (rsa) layer works the best. In the rest of the experiments, we continue with 2-layer rsa as the classifier-transfer method.

\subsubsection{Temporal Adaptation}

\begin{table*}
\begin{center}
\caption{Comparison of temporal-adaptation techniques for three different affine classifier-transfer models. Majority-voting is an image-to-image baseline (image-level classifier transfer). $\dagger$ are cases where the IMDB classifier is fine-tuned with WildestFaces training set.}
\label{fig:AffineComp} 
\label{avgpoolvsatp}
\begin{tabular}{lccccccccc|}
\toprule
&\multicolumn{2}{c}{None} &\multicolumn{2}{c}{1-layer affine} &\multicolumn{2}{c}{2-layer affine} &\multicolumn{2}{c}{2-layer res affine (rsa)} \\
\cmidrule{2-3}\cmidrule{4-5}\cmidrule{6-7}\cmidrule{8-9}
& val-h & test-h & val-h & test-h & val-h & test-h & val-h & test-h\\
\midrule
Maj. Voting & 30.5 & 24.6 & 37.7 & 28.8 & 31.4 & 26.3 & 43.11 & 31.9\\
AvgPool & 30.6 & 27.3 & 43.6 & \textbf{35.4} & 45.7 & \textbf{35.4} & \textbf{46.6} & 34.9\\
ATP (ours) & 36.5 & 32.6 & 44.8 & 39.3 & 42.9 & 35.3 & \textbf{47.4} & \textbf{39.3}\\
\midrule
Maj. Voting$\dagger$ &38.1&30.0&39.3&30.0&37.1&30.3&46.2&34.6 \\
AvgPool$\dagger$ & 40.6 & 32.7 & 45.2 & 35.3 & 43.6 & 36.5 &\textbf{48.4} & \textbf{40.5} \\
ATP (ours)$\dagger$ & 43.0 & 35.3 & 45.1  & 36.0 & 49.8 & 42.2 & \textbf{51.8} & \textbf{45.8}\\
\bottomrule
\end{tabular}
\vspace{-5mm}
\end{center}
\end{table*}

\begin{table*}
\centering{}\caption{Comparison of temporal-adaptation techniques for video representation.
We report the separate accuracies of seen and unseen classes, together
with their harmonic mean. $\dagger$ represents the case that the IMDB classifier is fine-tuned with WildestFaces training set.}
\label{tab:videoPoolingResults} %
\begin{tabular}{lccccccc}
\toprule
 & \multicolumn{3}{c}{{w/o classifier-transfer}} & \hspace{.5cm} & \multicolumn{3}{c}{w/ classifier-transfer}\tabularnewline
\cmidrule{2-8}

                                            & {seen } & {unseen } & {harmonic} &  & seen & unseen & harmonic\tabularnewline
\midrule
AvgPool                                 &   29.3              &  25.5                 & 27.3            &  & 36.7     &  33.2      & 34.9 \tabularnewline
DAN~\cite{rao2017learning}                  &  5.0               &   2.9                & 3.7                &  &  5.2    &    6.5  & 5.6 \tabularnewline
Self-attention \cite{vaswani2017attention}                       &   \textbf {37.2}             &     \textbf{ 34.5}             & \textbf{35.8}             &  &  37.1    & 34.7        & 35.9  \tabularnewline
ATP (ours)                         &    35.3         &      30.3         &  32.6         &  & 41.6 &
 37.3 & 39.3  \tabularnewline 
ATP (ours) $\dagger$        &    34.5         &      31.2         &  32.7         &  & \textbf{47.1} &
 \textbf{44.6} & \textbf{45.8}  \tabularnewline 
\bottomrule
\end{tabular}
    \vspace{-2mm}
\end{table*}


In Table \ref{avgpoolvsatp}, average pooling, majority voting and ATP are evaluated. Compared to vanilla AvgPool, vanilla ATP increases accuracy more than 5 points. ATP with 2-layer residual affine layer increases val-h and test-h even further, from 30.6 and 27.3 to 47.4 and 39.3, respectively. Finetuning the IMDB classifier with WildestFaces training set increases the accuracy by another 5 points to 45.8.

\begin{figure}
\begin{center}
\setlength{\tabcolsep}{3pt} 
        \begin{tabular}{c|cccc|c}
         \toprule
                \textbf{IMDB} & \multicolumn{4}{c|}{\textbf{Original Face Tracks}} & \textbf{DAN } \\
               
                \includegraphics[width=1.1cm,height=1.1cm]{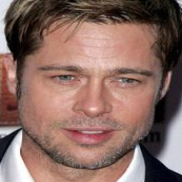} &
                 \includegraphics[width=1.1cm,height=1.1cm]{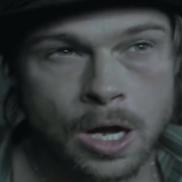} &
                 \includegraphics[width=1.1cm,height=1.1cm]{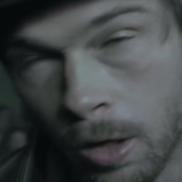} &
                 \includegraphics[width=1.1cm,height=1.1cm]{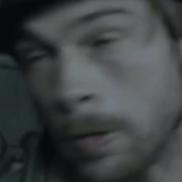} &
                \includegraphics[width=1.1cm,height=1.1cm]{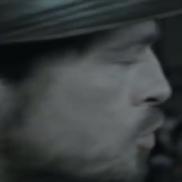} &
                \includegraphics[width=1.1cm,height=1.1cm]{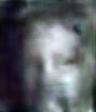} 
                
                \\
                
                 \includegraphics[width=1.1cm,height=1.1cm]{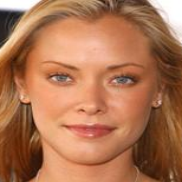} &
                 \includegraphics[width=1.1cm,height=1.1cm]{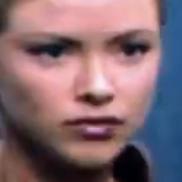} &
                 \includegraphics[width=1.1cm,height=1.1cm]{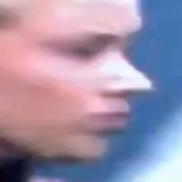} &
                 \includegraphics[width=1.1cm,height=1.1cm]{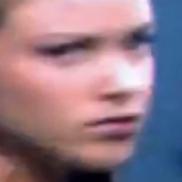} &
                 \includegraphics[width=1.1cm,height=1.1cm]{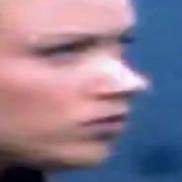}  &
                 \includegraphics[width=1.1cm,height=1.1cm]{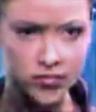}
                
                \\
                 
                \includegraphics[width=1.1cm,height=1.1cm]{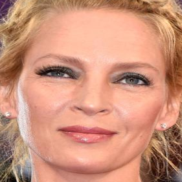} &
                 \includegraphics[width=1.1cm,height=1.1cm]{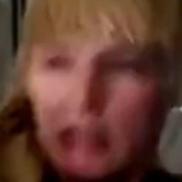} &
                 \includegraphics[width=1.1cm,height=1.1cm]{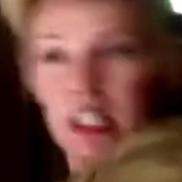} &
                 \includegraphics[width=1.1cm,height=1.1cm]{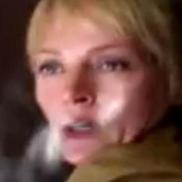}  &
                \includegraphics[width=1.1cm,height=1.1cm]{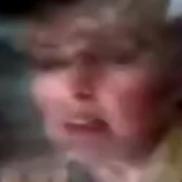}  &
                \includegraphics[width=1.1cm,height=1.1cm]{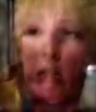}
                
                \\
                 
                \includegraphics[width=1.1cm,height=1.1cm]{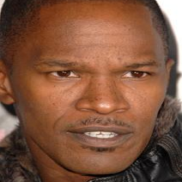} &
                 \includegraphics[width=1.1cm,height=1.1cm]{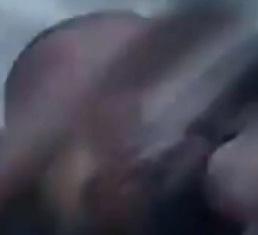} &
                 \includegraphics[width=1.1cm,height=1.1cm]{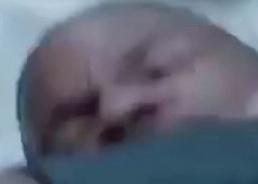} &
                 \includegraphics[width=1.1cm,height=1.1cm]{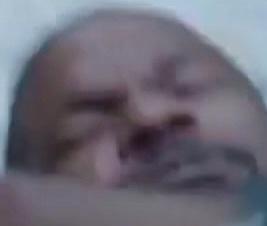}  &
                \includegraphics[width=1.1cm,height=1.1cm]{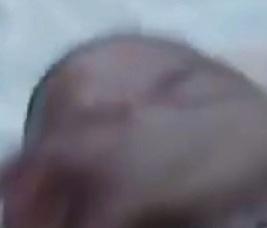}  &
                \includegraphics[width=1.1cm,height=1.1cm]{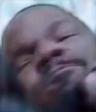}

                \\
                 
                \includegraphics[width=1.1cm,height=1.1cm]{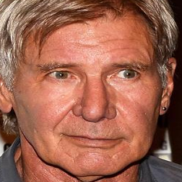} &
                 \includegraphics[width=1.1cm,height=1.1cm]{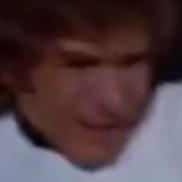} &
                 \includegraphics[width=1.1cm,height=1.1cm]{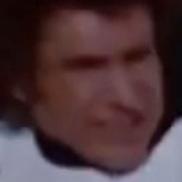} &
                 \includegraphics[width=1.1cm,height=1.1cm]{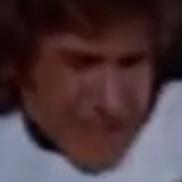} &
                 \includegraphics[width=1.1cm,height=1.1cm]{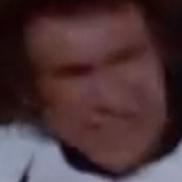} &
               \includegraphics[width=1.1cm,height=1.1cm]{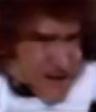} 
                
                \\
                 
                \includegraphics[width=1.1cm,height=1.1cm]{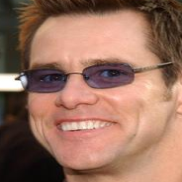} &
                 \includegraphics[width=1.1cm,height=1.1cm]{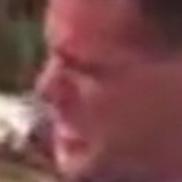} &
                 \includegraphics[width=1.1cm,height=1.1cm]{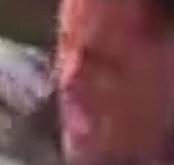} &
                 \includegraphics[width=1.1cm,height=1.1cm]{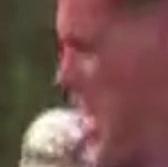} &
                 \includegraphics[width=1.1cm,height=1.1cm]{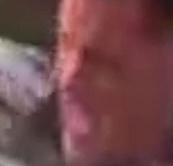} &
               \includegraphics[width=1.1cm,height=1.1cm]{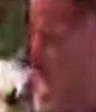}

                \\
                
                \includegraphics[width=1.1cm,height=1.1cm]{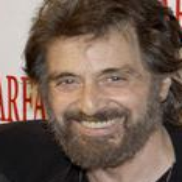} &
                 \includegraphics[width=1.1cm,height=1.1cm]{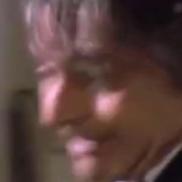} &
                 \includegraphics[width=1.1cm,height=1.1cm]{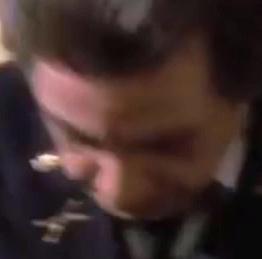} &
                 \includegraphics[width=1.1cm,height=1.1cm]{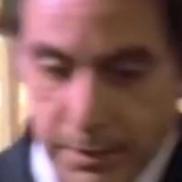} &
               \includegraphics[width=1.1cm,height=1.1cm]{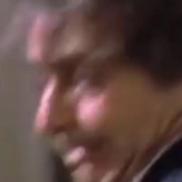} &
                 \includegraphics[width=1.1cm,height=1.1cm]{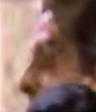}

                \\
                
                \includegraphics[width=1.1cm,height=1.1cm]{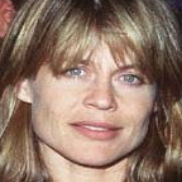} &
                 \includegraphics[width=1.1cm,height=1.1cm]{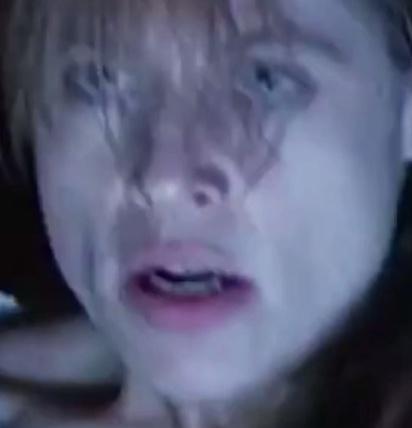} &
                 \includegraphics[width=1.1cm,height=1.1cm]{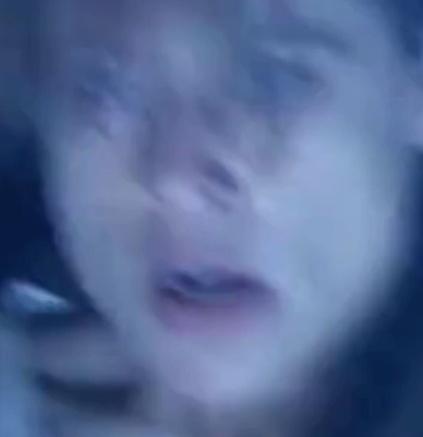} &
                 \includegraphics[width=1.1cm,height=1.1cm]{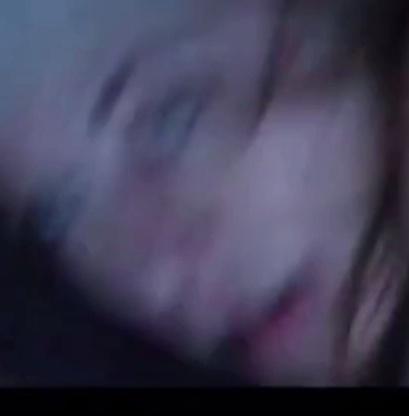} &
               \includegraphics[width=1.1cm,height=1.1cm]{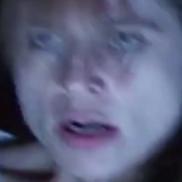} &
                 \includegraphics[width=1.1cm,height=1.1cm]{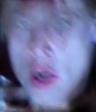}
                \\
                 \bottomrule
        \end{tabular}
    \caption{Example results for ATP with classifier-transfer. The first and last 4 rows depict examples for correct and incorrect classifications, respectively.
    The corresponding DAN \cite{rao2017learning} generated images (the rightmost column) are mostly noisy and imprecise.}
    \label{fig:qualitativeResults}
    \vspace{-8mm}
\end{center}
\end{figure}

We compare our proposed method with another aggregation method DAN~\cite{rao2017learning}, which is amongst the state-of-the-art methods for video face recognition. DAN ~\cite{rao2017learning} aggregates the information of an input video into one or few discriminative image(s) by using a GAN-based approach. For each face sequence, we generate an image using DAN model pre-trained on Youtube Faces(YTF)~\cite{wolf2011face} dataset. Example images generated by DAN~\cite{rao2017learning} is shown in Figure \ref{fig:qualitativeResults}. Inevitably, the images generated by this GAN-based model are not precise, due to noisy input sequences (Figure \ref{fig:qualitativeResults}).

DAN \cite{rao2017learning} fails to extend to different domains and  unseen classes (see Table \ref{tab:videoPoolingResults}). Self-attention \cite{vaswani2017attention} adaptation performs well yet enjoys slight improvements with classifier transfer. We argue this is due to its implicit classifier transfer mechanism (multi-head attention) as its high complexity can be harmful in our data-sparse setting. Ultimately, ATP outperforms other baselines with a clear margin.

\vspace{-2mm}
\section{Conclusion} \label{conclude}

In common surveillance scenarios, one may only have access to a clean photo of a person but may need to recognize the person in an unconstrained setting. In line with such scenarios, we study the partially-supervised domain-transfer problem within the context of face recognition, where algorithms are evaluated for their ability to recognize people in videos with violence, based on clean train images. 

We introduce the WildestFaces dataset that contains adverse effects at their extreme, such as blur, pose diversity, occlusions and resolution, and a principled evaluation protocol. Towards tackling the partially-supervised domain transfer, we propose (i) affine layers for classifier transfer, and, (ii) attention-based pooling for temporal adaptation. Compared to a number of strong baselines, including a self-attention based model, we show the proposed techniques outperform the baselines. We also highlight the challenges of this newly introduced dataset and the problem definition.

\textbf{Acknowledgments.} This work was supported in part by the TUBITAK Grant 116E445. We thank Oguzhan Oguz for his help in collecting and annotating the dataset.

{\small
\bibliographystyle{ieee_fullname}
\bibliography{egbib}

\begin{thebibliography}{10}\itemsep=-1pt

\bibitem{ahonen2006face}
Timo Ahonen, Abdenour Hadid, and Matti Pietikainen.
\newblock Face description with local binary patterns: Application to face
  recognition.
\newblock {\em IEEE Trans. Pattern Anal. Mach. Intell.}, 28(12):2037--2041,
  2006.

\bibitem{beveridge2013challenge}
J~Ross Beveridge, P~Jonathon Phillips, David~S Bolme, Bruce~A Draper, Geof~H
  Givens, Yui~Man Lui, Mohammad~Nayeem Teli, Hao Zhang, W~Todd Scruggs, Kevin~W
  Bowyer, et~al.
\newblock The challenge of face recognition from digital point-and-shoot
  cameras.
\newblock In {\em IEEE International Conference on Biometrics}, pages 1--8,
  2013.

\bibitem{bousmalis2016domain}
Konstantinos Bousmalis, George Trigeorgis, Nathan Silberman, Dilip Krishnan,
  and Dumitru Erhan.
\newblock Domain separation networks.
\newblock In {\em Proc. Adv. Neural Inf. Process. Syst.}, pages 343--351, 2016.

\bibitem{cao2017vggface2}
Qiong Cao, Li Shen, Weidi Xie, Omkar~M Parkhi, and Andrew Zisserman.
\newblock Vggface2: A dataset for recognising faces across pose and age.
\newblock In {\em 2018 13th IEEE International Conference on Automatic Face \&
  Gesture Recognition (FG 2018)}, pages 67--74. IEEE, 2018.

\bibitem{chao16gzsl}
Wei-Lun Chao, Soravit Changpinyo, Boqing Gong, and Fei Sha.
\newblock An empirical study and analysis of generalized zero-shot learning for
  object recognition in the wild.
\newblock In {\em European Conference on Computer Vision}, pages 52--68.
  Springer, 2016.

\bibitem{chen2017show}
Tseng-Hung Chen, Yuan-Hong Liao, Ching-Yao Chuang, Wan~Ting Hsu, Jianlong Fu,
  and Min Sun.
\newblock Show, adapt and tell: Adversarial training of cross-domain image
  captioner.
\newblock In {\em Proc. IEEE Int. Conf. on Computer Vision}, pages 521--530,
  2017.

\bibitem{cheng2018duplex}
Gong Cheng, Peicheng Zhou, and Junwei Han.
\newblock Duplex metric learning for image set classification.
\newblock {\em IEEE Trans. on Image Processing}, 27(1):281--292, 2018.

\bibitem{chowdhury2016one}
Aruni~Roy Chowdhury, Tsung-Yu Lin, Subhransu Maji, and Erik Learned-Miller.
\newblock One-to-many face recognition with bilinear cnns.
\newblock In {\em IEEE Winter Conf. on Appl. of Comput. Vis.}, pages 1--9,
  2016.

\bibitem{coelho2018face}
Daniel Coelho~de Castro and Sebastian Nowozin.
\newblock From face recognition to models of identity: A bayesian approach to
  learning about unknown identities from unsupervised data.
\newblock In {\em Proceedings of the European Conference on Computer Vision
  (ECCV)}, pages 745--761, 2018.

\bibitem{csurka2017domain}
Gabriela Csurka.
\newblock Domain adaptation for visual applications: {A} comprehensive survey.
\newblock {\em CoRR}, abs/1702.05374, 2017.

\bibitem{csurka2016unsupervised}
Gabriela Csurka, Boris Chidlowskii, St{\'e}phane Clinchant, and Sophia Michel.
\newblock Unsupervised domain adaptation with regularized domain instance
  denoising.
\newblock In {\em European Conference on Computer Vision}, pages 458--466,
  2016.

\bibitem{deng2019arcface}
Jiankang Deng, Jia Guo, Niannan Xue, and Stefanos Zafeiriou.
\newblock Arcface: Additive angular margin loss for deep face recognition.
\newblock In {\em Proceedings of the IEEE Conference on Computer Vision and
  Pattern Recognition}, pages 4690--4699, 2019.

\bibitem{ding2016comprehensive}
Changxing Ding and Dacheng Tao.
\newblock A comprehensive survey on pose-invariant face recognition.
\newblock {\em ACM Transactions on intelligent systems and technology (TIST)},
  7(3):37, 2016.

\bibitem{edwards1998face}
Gareth~J Edwards, Timothy~F Cootes, and Christopher~J Taylor.
\newblock Face recognition using active appearance models.
\newblock In {\em European conference on computer vision}, pages 581--595,
  1998.

\bibitem{ferrari2018extended}
Claudio Ferrari, Stefano Berretti, and Alberrto Del~Bimbo.
\newblock Extended youtube faces: a dataset for heterogeneous open-set face
  identification.
\newblock In {\em 2018 24th International Conference on Pattern Recognition
  (ICPR)}, pages 3408--3413. IEEE, 2018.

\bibitem{ferrari2019discovering}
Claudio Ferrari, Stefano Berretti, and Alberto Del~Bimbo.
\newblock Discovering identity specific activation patterns in deep descriptors
  for template based face recognition.
\newblock In {\em 2019 14th IEEE International Conference on Automatic Face \&
  Gesture Recognition (FG 2019)}, pages 1--5. IEEE, 2019.

\bibitem{ganin2016domain}
Yaroslav Ganin, Evgeniya Ustinova, Hana Ajakan, Pascal Germain, Hugo
  Larochelle, Fran{\c{c}}ois Laviolette, Mario Marchand, and Victor Lempitsky.
\newblock Domain-adversarial training of neural networks.
\newblock {\em The Journal of Machine Learning Research}, 17(1):2096--2030,
  2016.

\bibitem{gebru2017fine}
Timnit Gebru, Judy Hoffman, and Li Fei-Fei.
\newblock Fine-grained recognition in the wild: A multi-task domain adaptation
  approach.
\newblock In {\em Proc. IEEE Int. Conf. on Computer Vision}, pages 1358--1367,
  2017.

\bibitem{ghifary2016deep}
Muhammad Ghifary, W~Bastiaan Kleijn, Mengjie Zhang, David Balduzzi, and Wen Li.
\newblock Deep reconstruction-classification networks for unsupervised domain
  adaptation.
\newblock In {\em European Conference on Computer Vision}, pages 597--613,
  2016.

\bibitem{gidaris18gfew}
Spyros Gidaris and Nikos Komodakis.
\newblock Dynamic {Few}-{Shot} {Visual} {Learning} {Without} {Forgetting}.
\newblock In {\em Proc. IEEE Conf. Comput. Vis. Pattern Recog.}, 2018.

\bibitem{glorot2010understanding}
Xavier Glorot and Yoshua Bengio.
\newblock Understanding the difficulty of training deep feedforward neural
  networks.
\newblock In {\em Proceedings of the thirteenth international conference on
  artificial intelligence and statistics}, pages 249--256, 2010.

\bibitem{goodfellow2014generative}
Ian Goodfellow, Jean Pouget-Abadie, Mehdi Mirza, Bing Xu, David Warde-Farley,
  Sherjil Ozair, Aaron Courville, and Yoshua Bengio.
\newblock Generative adversarial nets.
\newblock In {\em Proc. Adv. Neural Inf. Process. Syst.}, pages 2672--2680,
  2014.

\bibitem{goswami2014mdlface}
Gaurav Goswami, Romil Bhardwaj, Richa Singh, and Mayank Vatsa.
\newblock Mdlface: Memorability augmented deep learning for video face
  recognition.
\newblock In {\em IEEE International Joint Conference on Biometrics}, pages
  1--7, 2014.

\bibitem{goswami2017face}
Gaurav Goswami, Mayank Vatsa, and Richa Singh.
\newblock Face verification via learned representation on feature-rich video
  frames.
\newblock {\em IEEE Transactions on Information Forensics and Security},
  12(7):1686--1698, 2017.

\bibitem{guo2016ms}
Yandong Guo, Lei Zhang, Yuxiao Hu, Xiaodong He, and Jianfeng Gao.
\newblock Ms-celeb-1m: Challenge of recognizing one million celebrities in the
  real world.
\newblock {\em Electronic Imaging}, 2016(11):1--6, 2016.

\bibitem{hassner2016pooling}
Tal Hassner, Iacopo Masi, Jungyeon Kim, Jongmoo Choi, Shai Harel, Prem
  Natarajan, and Gerard Medioni.
\newblock Pooling faces: Template based face recognition with pooled face
  images.
\newblock In {\em Proceedings of the IEEE conference on computer vision and
  pattern recognition workshops}, pages 59--67, 2016.

\bibitem{hu2019noise}
Wei Hu, Yangyu Huang, Fan Zhang, and Ruirui Li.
\newblock Noise-tolerant paradigm for training face recognition cnns.
\newblock In {\em Proceedings of the IEEE Conference on Computer Vision and
  Pattern Recognition}, pages 11887--11896, 2019.

\bibitem{huang2007labeled}
Gary~B Huang, Manu Ramesh, Tamara Berg, and Erik Learned-Miller.
\newblock Labeled faces in the wild: A database for studying face recognition
  in unconstrained environments.
\newblock Technical report, Technical Report 07-49, University of
  Massachusetts, Amherst, 2007.

\bibitem{huang2015benchmark}
Zhiwu Huang, Shiguang Shan, Ruiping Wang, Haihong Zhang, Shihong Lao, Alifu
  Kuerban, and Xilin Chen.
\newblock A benchmark and comparative study of video-based face recognition on
  cox face database.
\newblock {\em IEEE Trans. on Image Processing}, 24(12):5967--5981, 2015.

\bibitem{huang2015log}
Zhiwu Huang, Ruiping Wang, Shiguang Shan, Xianqiu Li, and Xilin Chen.
\newblock Log-euclidean metric learning on symmetric positive definite manifold
  with application to image set classification.
\newblock In {\em Proc. Int. Conf. Mach. Learn.}, pages 720--729, 2015.

\bibitem{huang2017cross}
Zhiwu Huang, Ruiping Wang, Luc Van~Gool, Xilin Chen, et~al.
\newblock Cross euclidean-to-riemannian metric learning with application to
  face recognition from video.
\newblock {\em IEEE Trans. Pattern Anal. Mach. Intell.}, 2017.

\bibitem{jain2010fddb}
Vidit Jain and Erik Learned-Miller.
\newblock Fddb: A benchmark for face detection in unconstrained settings.
\newblock {\em University of Massachusetts, Amherst, Tech. Rep.
  UM-CS-2010-009}, 2(7):8, 2010.

\bibitem{kalka2018ijb}
Nathan~D Kalka, Brianna Maze, James~A Duncan, Kevin O’Connor, Stephen
  Elliott, Kaleb Hebert, Julia Bryan, and Anil~K Jain.
\newblock Ijb--s: Iarpa janus surveillance video benchmark.
\newblock In {\em 2018 IEEE 9th International Conference on Biometrics Theory,
  Applications and Systems (BTAS)}, pages 1--9. IEEE, 2018.

\bibitem{kemelmacher2016megaface}
Ira Kemelmacher-Shlizerman, Steven~M Seitz, Daniel Miller, and Evan Brossard.
\newblock The megaface benchmark: 1 million faces for recognition at scale.
\newblock In {\em Proc. IEEE Conf. Comput. Vis. Pattern Recog.}, pages
  4873--4882, 2016.

\bibitem{klare2015pushing}
Brendan~F Klare, Ben Klein, Emma Taborsky, Austin Blanton, Jordan Cheney,
  Kristen Allen, Patrick Grother, Alan Mah, and Anil~K Jain.
\newblock Pushing the frontiers of unconstrained face detection and
  recognition: Iarpa janus benchmark a.
\newblock In {\em Proc. IEEE Conf. Comput. Vis. Pattern Recog.}, pages
  1931--1939, 2015.

\bibitem{kushwaha2018disguised}
Vineet Kushwaha, Maneet Singh, Richa Singh, Mayank Vatsa, Nalini Ratha, and
  Rama Chellappa.
\newblock Disguised faces in the wild.
\newblock In {\em Proceedings of the IEEE Conference on Computer Vision and
  Pattern Recognition Workshops}, pages 1--9, 2018.

\bibitem{li2013probabilistic}
Haoxiang Li, Gang Hua, Zhe Lin, Jonathan Brandt, and Jianchao Yang.
\newblock Probabilistic elastic matching for pose variant face verification.
\newblock In {\em Proc. IEEE Conf. Comput. Vis. Pattern Recog.}, pages
  3499--3506, 2013.

\bibitem{li2014eigen}
Haoxiang Li, Gang Hua, Xiaohui Shen, Zhe Lin, and Jonathan Brandt.
\newblock Eigen-pep for video face recognition.
\newblock In {\em Asian Conference on Computer Vision}, pages 17--33, 2014.

\bibitem{Long:2015:LTF:3045118.3045130}
Mingsheng Long, Yue Cao, Jianmin Wang, and Michael~I. Jordan.
\newblock Learning transferable features with deep adaptation networks.
\newblock In {\em Proceedings of the 32Nd International Conference on
  International Conference on Machine Learning - Volume 37}, ICML'15, pages
  97--105. JMLR.org, 2015.

\bibitem{long2014adaptation}
Mingsheng Long, Jianmin Wang, Guiguang Ding, Sinno~Jialin Pan, and S~Yu Philip.
\newblock Adaptation regularization: A general framework for transfer learning.
\newblock {\em IEEE Transactions on Knowledge and Data Engineering},
  26(5):1076--1089, 2014.

\bibitem{long2016unsupervised}
Mingsheng Long, Han Zhu, Jianmin Wang, and Michael~I Jordan.
\newblock Unsupervised domain adaptation with residual transfer networks.
\newblock In {\em Advances in Neural Information Processing Systems}, pages
  136--144, 2016.

\bibitem{nagrani2018benedict}
Arsha Nagrani and Andrew Zisserman.
\newblock From benedict cumberbatch to sherlock holmes: Character
  identification in tv series without a script.
\newblock In {\em British Machine Vision Conference}, 2017.

\bibitem{ng2014data}
Hong-Wei Ng and Stefan Winkler.
\newblock A data-driven approach to cleaning large face datasets.
\newblock In {\em IEEE International Conference on Image Processing}, pages
  343--347, 2014.

\bibitem{parkhi2014compact}
Omkar~M Parkhi, Karen Simonyan, Andrea Vedaldi, and Andrew Zisserman.
\newblock A compact and discriminative face track descriptor.
\newblock In {\em Proc. IEEE Conf. Comput. Vis. Pattern Recog.}, pages
  1693--1700, 2014.

\bibitem{parkhi2015deep}
Omkar~M Parkhi, Andrea Vedaldi, Andrew Zisserman, et~al.
\newblock Deep face recognition.
\newblock In {\em British Machine Vision Conference}, 2015.

\bibitem{paszke2017pytorch}
Adam Paszke, Sam Gross, Soumith Chintala, and Gregory Chanan.
\newblock Pytorch, 2017.

\bibitem{pech2000diatom}
Jos{\'e}~Luis Pech-Pacheco, Gabriel Crist{\'o}bal, Jes{\'u}s Chamorro-Martinez,
  and Joaqu{\'\i}n Fern{\'a}ndez-Valdivia.
\newblock Diatom autofocusing in brightfield microscopy: a comparative study.
\newblock In {\em IAPR International Conference on Pattern Recognition},
  volume~3, pages 314--317, 2000.

\bibitem{rao2017learning}
Yongming Rao, Ji Lin, Jiwen Lu, and Jie Zhou.
\newblock Learning discriminative aggregation network for video-based face
  recognition.
\newblock In {\em Proc. IEEE Conf. Comput. Vis. Pattern Recog.}, pages
  3781--3790, 2017.

\bibitem{rao2017attention}
Yongming Rao, Jiwen Lu, and Jie Zhou.
\newblock Attention-aware deep reinforcement learning for video face
  recognition.
\newblock In {\em Proc. IEEE Conf. Comput. Vis. Pattern Recog.}, pages
  3931--3940, 2017.

\bibitem{RotheIJCV2016}
Rasmus Rothe, Radu Timofte, and Luc~Van Gool.
\newblock Deep expectation of real and apparent age from a single image without
  facial landmarks.
\newblock {\em Int. J. on Computer Vision}, July 2016.

\bibitem{ruiz2017fine}
Nataniel Ruiz, Eunji Chong, and James~M Rehg.
\newblock Fine-grained head pose estimation without keypoints.
\newblock In {\em Proceedings of the IEEE Conference on Computer Vision and
  Pattern Recognition Workshops}, pages 2074--2083, 2018.

\bibitem{schroff2015facenet}
Florian Schroff, Dmitry Kalenichenko, and James Philbin.
\newblock Facenet: A unified embedding for face recognition and clustering.
\newblock In {\em Proc. IEEE Conf. Comput. Vis. Pattern Recog.}, pages
  815--823, 2015.

\bibitem{sengupta2016frontal}
Soumyadip Sengupta, Jun-Cheng Chen, Carlos Castillo, Vishal~M Patel, Rama
  Chellappa, and David~W Jacobs.
\newblock Frontal to profile face verification in the wild.
\newblock In {\em 2016 IEEE Winter Conference on Applications of Computer
  Vision (WACV)}, pages 1--9. IEEE, 2016.

\bibitem{sun2016return}
Baochen Sun, Jiashi Feng, and Kate Saenko.
\newblock Return of frustratingly easy domain adaptation.
\newblock In {\em AAAI}, 2016.

\bibitem{sun2014deep}
Yi Sun, Yuheng Chen, Xiaogang Wang, and Xiaoou Tang.
\newblock Deep learning face representation by joint
  identification-verification.
\newblock In {\em Proc. Adv. Neural Inf. Process. Syst.}, pages 1988--1996,
  2014.

\bibitem{sun2015deepid3}
Yi Sun, Ding Liang, Xiaogang Wang, and Xiaoou Tang.
\newblock Deepid3: Face recognition with very deep neural networks.
\newblock {\em arXiv preprint arXiv:1502.00873}, 2015.

\bibitem{sun2013hybrid}
Yi Sun, Xiaogang Wang, and Xiaoou Tang.
\newblock Hybrid deep learning for face verification.
\newblock In {\em Proc. IEEE Int. Conf. on Computer Vision}, pages 1489--1496,
  2013.

\bibitem{taigman2014deepface}
Yaniv Taigman, Ming Yang, Marc'Aurelio Ranzato, and Lior Wolf.
\newblock Deepface: Closing the gap to human-level performance in face
  verification.
\newblock In {\em Proc. IEEE Conf. Comput. Vis. Pattern Recog.}, pages
  1701--1708, 2014.

\bibitem{tran2017disentangled}
Luan Tran, Xi Yin, and Xiaoming Liu.
\newblock Disentangled representation learning gan for pose-invariant face
  recognition.
\newblock In {\em Proc. IEEE Conf. Comput. Vis. Pattern Recog.}, 2017.

\bibitem{turk1991face}
Matthew~A Turk and Alex~P Pentland.
\newblock Face recognition using eigenfaces.
\newblock In {\em Proc. IEEE Conf. Comput. Vis. Pattern Recog.}, pages
  586--591, 1991.

\bibitem{tzeng2015simultaneous}
Eric Tzeng, Judy Hoffman, Trevor Darrell, and Kate Saenko.
\newblock Simultaneous deep transfer across domains and tasks.
\newblock In {\em Proc. IEEE Int. Conf. on Computer Vision}, pages 4068--4076,
  2015.

\bibitem{tzeng2017adversarial}
Eric Tzeng, Judy Hoffman, Kate Saenko, and Trevor Darrell.
\newblock Adversarial discriminative domain adaptation.
\newblock In {\em Proceedings of the IEEE Conference on Computer Vision and
  Pattern Recognition}, pages 7167--7176, 2017.

\bibitem{vaswani2017attention}
Ashish Vaswani, Noam Shazeer, Niki Parmar, Jakob Uszkoreit, Llion Jones,
  Aidan~N Gomez, {\L}ukasz Kaiser, and Illia Polosukhin.
\newblock Attention is all you need.
\newblock In {\em Proc. Adv. Neural Inf. Process. Syst.}, pages 5998--6008,
  2017.

\bibitem{wang2018cosface}
Hao Wang, Yitong Wang, Zheng Zhou, Xing Ji, Dihong Gong, Jingchao Zhou, Zhifeng
  Li, and Wei Liu.
\newblock Cosface: Large margin cosine loss for deep face recognition.
\newblock In {\em Proceedings of the IEEE Conference on Computer Vision and
  Pattern Recognition}, pages 5265--5274, 2018.

\bibitem{wang2017deep}
Yifei Wang, Wen Li, Dengxin Dai, and Luc~Van Gool.
\newblock Deep domain adaptation by geodesic distance minimization.
\newblock In {\em 2017 IEEE International Conference on Computer Vision
  Workshops (ICCVW)}, 2017.

\bibitem{wen2016discriminative}
Yandong Wen, Kaipeng Zhang, Zhifeng Li, and Yu Qiao.
\newblock A discriminative feature learning approach for deep face recognition.
\newblock In {\em European Conference on Computer Vision}, pages 499--515,
  2016.

\bibitem{wiskott1997face}
Laurenz Wiskott, Norbert Kr{\"u}ger, N Kuiger, and Christoph Von Der~Malsburg.
\newblock Face recognition by elastic bunch graph matching.
\newblock {\em IEEE Trans. Pattern Anal. Mach. Intell.}, 19(7):775--779, 1997.

\bibitem{wolf2011face}
Lior Wolf, Tal Hassner, and Itay Maoz.
\newblock Face recognition in unconstrained videos with matched background
  similarity.
\newblock In {\em Proc. IEEE Conf. Comput. Vis. Pattern Recog.}, pages
  529--534, 2011.

\bibitem{wright2009robust}
John Wright, Allen~Y Yang, Arvind Ganesh, S~Shankar Sastry, and Yi Ma.
\newblock Robust face recognition via sparse representation.
\newblock {\em IEEE Trans. Pattern Anal. Mach. Intell.}, 31(2):210--227, 2009.

\bibitem{wu2017compact}
Chunpeng Wu, Wei Wen, Tariq Afzal, Yongmei Zhang, Yiran Chen, and Hai Li.
\newblock A compact {DNN:} approaching googlenet-level accuracy of
  classification and domain adaptation.
\newblock In {\em Proc. IEEE Conf. Comput. Vis. Pattern Recog.}, 2017.

\bibitem{wu2018light}
Xiang Wu, Ran He, Zhenan Sun, and Tieniu Tan.
\newblock A light cnn for deep face representation with noisy labels.
\newblock {\em IEEE Transactions on Information Forensics and Security},
  13(11):2884--2896, 2018.

\bibitem{wulfmeier2017addressing}
Markus Wulfmeier, Alex Bewley, and Ingmar Posner.
\newblock Addressing appearance change in outdoor robotics with adversarial
  domain adaptation.
\newblock In {\em IEEE/RSJ International Conference on Intelligent Robots and
  Systems (IROS)}, 2017.

\bibitem{xian2017zero}
Yongqin Xian, Bernt Schiele, and Zeynep Akata.
\newblock Zero-shot learning-the good, the bad and the ugly.
\newblock In {\em Proc. IEEE Conf. Comput. Vis. Pattern Recog.}, pages
  4582--4591, 2017.

\bibitem{xie2010fusing}
Shufu Xie, Shiguang Shan, Xilin Chen, and Jie Chen.
\newblock Fusing local patterns of gabor magnitude and phase for face
  recognition.
\newblock {\em IEEE Trans. on Image Processing}, 19(5):1349--1361, 2010.

\bibitem{xie2018multicolumn}
Weidi Xie and Andrew Zisserman.
\newblock Multicolumn networks for face recognition.
\newblock In {\em British Machine Vision Conference}, 2018.

\bibitem{yan2017mind}
Hongliang Yan, Yukang Ding, Peihua Li, Qilong Wang, Yong Xu, and Wangmeng Zuo.
\newblock Mind the class weight bias: Weighted maximum mean discrepancy for
  unsupervised domain adaptation.
\newblock In {\em Proc. IEEE Conf. Comput. Vis. Pattern Recog.}, volume~3,
  2017.

\bibitem{yan2014face}
Junjie Yan, Xuzong Zhang, Zhen Lei, and Stan~Z Li.
\newblock Face detection by structural models.
\newblock {\em Image and Vision Computing}, 32(10):790--799, 2014.

\bibitem{yang2017neural}
Jiaolong Yang, Peiran Ren, Dongqing Zhang, Dong Chen, Fang Wen, Hongdong Li,
  and Gang Hua.
\newblock Neural aggregation network for video face recognition.
\newblock In {\em Proceedings of the IEEE Conference on Computer Vision and
  Pattern Recognition}, pages 4362--4371, 2017.

\bibitem{yang2016wider}
Shuo Yang, Ping Luo, Chen-Change Loy, and Xiaoou Tang.
\newblock Wider face: A face detection benchmark.
\newblock In {\em Proc. IEEE Conf. Comput. Vis. Pattern Recog.}, pages
  5525--5533, 2016.

\bibitem{yin2019feature}
Xi Yin, Xiang Yu, Kihyuk Sohn, Xiaoming Liu, and Manmohan Chandraker.
\newblock Feature transfer learning for face recognition with under-represented
  data.
\newblock In {\em Proceedings of the IEEE Conference on Computer Vision and
  Pattern Recognition}, pages 5704--5713, 2019.

\bibitem{zhan2018consensus}
Xiaohang Zhan, Ziwei Liu, Junjie Yan, Dahua Lin, and Chen Change~Loy.
\newblock Consensus-driven propagation in massive unlabeled data for face
  recognition.
\newblock In {\em Proceedings of the European Conference on Computer Vision
  (ECCV)}, pages 568--583, 2018.

\bibitem{zhao2017dual}
Jian Zhao, Lin Xiong, Panasonic~Karlekar Jayashree, Jianshu Li, Fang Zhao,
  Zhecan Wang, Panasonic~Sugiri Pranata, Panasonic~Shengmei Shen, Shuicheng
  Yan, and Jiashi Feng.
\newblock Dual-agent gans for photorealistic and identity preserving profile
  face synthesis.
\newblock In {\em Advances in Neural Information Processing Systems}, pages
  66--76, 2017.

\bibitem{zheng2018ring}
Yutong Zheng, Dipan~K Pal, and Marios Savvides.
\newblock Ring loss: Convex feature normalization for face recognition.
\newblock In {\em Proceedings of the IEEE conference on computer vision and
  pattern recognition}, pages 5089--5097, 2018.

\bibitem{zhong2018ghostvlad}
Yujie Zhong, Relja Arandjelovi{\'c}, and Andrew Zisserman.
\newblock Ghostvlad for set-based face recognition.
\newblock In {\em Asian Conf. on Computer Vision}, 2018.

\bibitem{zhu2012face}
Xiangxin Zhu and Deva Ramanan.
\newblock Face detection, pose estimation, and landmark localization in the
  wild.
\newblock In {\em Proc. IEEE Conf. Comput. Vis. Pattern Recog.}, 2012.

\end{thebibliography}
}

\end{document}